\documentclass[conference]{IEEEtran}
\IEEEoverridecommandlockouts
\usepackage{cite}
\usepackage{amsmath,amssymb,amsfonts}
\usepackage{algorithmic}
\usepackage{graphicx}
\usepackage{textcomp}
\usepackage{xcolor}
\usepackage{float}
\usepackage{booktabs}
\usepackage{url}
\def\BibTeX{{\rm B\kern-.05em{\sc i\kern-.025em b}\kern-.08em
    T\kern-.1667em\lower.7ex\hbox{E}\kern-.125emX}}

\begin{document}

\title{Detail in Context: A Dual-Scale Machine Learning Framework for Mycosis Fungoides Detection
}

\author{
    \IEEEauthorblockN{Mohamed Hazem\IEEEauthorrefmark{1},
    Tarek Waleed\IEEEauthorrefmark{1},
    Omar Khaled\IEEEauthorrefmark{1},
    Nada Omar\IEEEauthorrefmark{1},
    Mahmoud Raslan\IEEEauthorrefmark{1},
    Marwa Mohamed Fawzy\IEEEauthorrefmark{2},}
    \IEEEauthorblockN{Aya Fahim\IEEEauthorrefmark{2},
    Rania M. Mogawer\IEEEauthorrefmark{2},
    Ahmed Mourad\IEEEauthorrefmark{2},
    Kariman Mansour\IEEEauthorrefmark{2},
    Muhammad Rushdi\IEEEauthorrefmark{1}}
    \vspace{0.15cm}
    \IEEEauthorblockA{\IEEEauthorrefmark{1}Faculty of Engineering, Cairo University, Giza, Egypt}
    \IEEEauthorblockA{\IEEEauthorrefmark{2}Dermatology Department, Faculty of Medicine, Cairo University, Cairo, Egypt}
}

\maketitle

\begin{abstract}
    Mycosis fungoides (MF) is a rare form of cutaneous T-cell lymphoma that is often misdiagnosed 
    in early stages due to its visual similarity to benign inflammatory dermatoses. 
    Early and accurate diagnosis is critical for improving patient outcomes. 
    In this paper, we propose a comprehensive diagnostic framework for automated MF detection 
    that combines dual-scale histopathological image analysis with deep learning. 
    To distinguish MF from other lymphoproliferative skin conditions, the proposed approach leverages 
    a late-fusion ensemble of dual-magnification (10$\times$ and 20$\times$) convolutional neural networks (CNNs), 
    complemented by a random forest classifier trained on 16 clinical features. 
    Experimental results on an expanded dataset of 6,267 images (4,306 MF; 1,961 Non-MF) across 463 patients 
    demonstrate that strong detection performance is obtained by prioritizing higher-resolution cytological details 
    (20$\times$) within broader architectural context (10$\times$). The image-based late-fusion model achieves an accuracy of 83.58\% and a sensitivity of 89.13\%, 
    while the clinical random forest model achieves an accuracy of 96.6\% and sensitivity of 93.8\%, 
    highlighting the potential of this multimodal framework as a robust clinical decision support system in dermatology.
    This framework addresses two distinct clinical objectives: an image-based 
    dual-scale pipeline optimized for the early diagnostic screening of MF versus non-MF 
    dermatoses, and a complementary clinical metadata model designed for the subsequent 
    staging of confirmed MF cases (patch/plaque versus tumor)
\end{abstract}

\section{Introduction}
Mycosis fungoides (MF) is a type of skin lymphoma characterized by malignant proliferation of skin-homing T
lymphocytes and is considered the most common primary cutaneous lymphoma (CL), making up 
almost 50\% of all cases \cite{willemze2005}. Despite its prevalence among cutaneous
T-cell lymphomas (CTCLs), MF remains a rare disease with an estimated incidence of 0.5 cases per 
100,000 person-years \cite{willemze2005}.
The clinical presentation of MF is highly variable, often manifesting as erythematous patches, 
plaques, or tumors that can easily be mistaken for
benign inflammatory skin conditions such as eczema or psoriasis. This diagnostic ambiguity contributes to significant delays in accurate diagnosis,
with studies reporting an average time to diagnosis of 3-6 years from symptom onset \cite{beatty2025_early_mf_diagnosis}.
MF remains challenging to diagnose, particularly in early stages, where clinical presentations often mimic benign skin disorders such as eczema or psoriasis.

Recent advances in deep learning have shown significant promise in medical image
analysis \cite{litjens2017deep}, particularly in dermatology \cite{esteva2017dermatologist}.
However, most existing studies
focus on common types of skin cancer such as melanoma, while rare diseases like MF remain 
underexplored. Our dual-scale image pipeline 
serves as an initial triage tool for early MF detection; while the clinical metadata 
classifier acts as a secondary mechanism to predict disease progression and staging in 
established cases, reflecting the sequential nature of dermatological workflows. 
This work aims to address this gap by developing and evaluating a
deep-learning-based system for MF detection in histopathological skin images.

The main contributions of this work can be summarized as follows:
\begin{itemize}
    \item A dual-scale MF modeling approach leveraging TF-EfficientNet-B3 architectures trained on histopathological images at 10$\times$ and 20$\times$ magnifications.
    \item A patient-level MF detection strategy based on mean aggregation of patch-level probabilities.
    \item A weighted late-fusion model integrating multi-magnification decisions that explicitly prioritizes cytological cellular details over low-power architectural context to maximize sensitivity.
    \item Robust machine-learning-based MF detectors trained on 16 complementary clinical features of 
    patient metadata, with a rigorous comparison between XGBoost, random forests, and logistic regression.
\end{itemize}

\section{Related Work}
Previous research in dermatological image analysis has primarily focused on melanoma detection and benign-versus-malignant lesion classification.
Studies employing convolutional neural networks (CNNs) have demonstrated dermatologist-level performance in some tasks \cite{Doeleman2024-MF}.
However, literature addressing MF specifically remains limited, largely due to data scarcity, subtlety of 
histopathological features,
and high inter-observer variability.

Recent works have explored AI-based approaches for lymphoproliferative skin diseases, yet MF is often excluded or significantly underrepresented, underscoring the
need for targeted investigation into AI-assisted MF diagnosis.

Almost all recent studies rely on digitized whole-slide images (WSIs) for histopathological analysis. In contrast, to address the lack of digital slide
scanners prevalent in resource-limited countries, our framework operates on conventional microscopic images acquired at fixed magnifications rather than WSIs.
This setting reflects real-world clinical constraints and represents a novel direction that is largely absent from recent literature.
Contemporary approaches predominantly employ weakly supervised deep learning frameworks, such as clustering-constrained attention multiple instance
learning (CLAM) \cite{Lu2021} or toolkits such as Trident \cite{zhang2025standardizing,vaidya2025molecular}, which are designed to leverage WSIs.
Our approach diverges from these paradigms by focusing on the analysis of optically magnified images under practical acquisition conditions.

\section{Materials and Methods}

\subsection{Dataset}
The dataset used in this study was collected through a clinical collaboration with the 
dermatopathology unit of a major tertiary care university hospital.
It consists of retrospective histopathological images captured from regions of interest with 
salient histopathological features discernible to a junior dermatologist, as well as associated clinical metadata.
Images were captured using a conventional microscope at two magnification levels of 10$\times$ and 20$\times$. 
These magnification levels refer specifically to the power of the objective lens (located near the slide). 
Because the microscope utilizes a fixed 10$\times$ eyepiece lens (located near the camera/observer), these objective magnifications 
correspond to total optical magnifications of 100$\times$ and 200$\times$, respectively. 
The dataset comprises 6,257 images from 463 patients (with 2146 and 4121 images captured respectively at magnification levels of 
10$\times$ and 20$\times$) reflecting real-world diagnostic challenges encountered in clinical practice.

\begin{table}[htbp]
    \centering
    \caption{Distribution of patients and images across diagnostic categories and magnifications.}
    \label{tab:dataset_distribution}
    \begin{tabular}{@{}lcc@{}}
        \toprule
        \textbf{Diagnostic Category} & \textbf{No. of Patients} & \textbf{No. of Images} \\ 
        \midrule
        \textbf{Mycosis Fungoides (MF)} & \textbf{311} & \textbf{4,306} \\
        \midrule
        \textbf{Non-MF Cohort (Total)} & \textbf{152} & \textbf{1,961} \\
        \hspace{4mm} B-cell lymphoma & 18 & 358 \\
        \hspace{4mm} PLEVA / PLC & 108 & 1,268 \\
        \hspace{4mm} T-cell dyscrasia & 15 & 180 \\
        \hspace{4mm} Pseudolymphoma & 11 & 155 \\
        \midrule
        \textbf{Total Dataset} & \textbf{463} & \textbf{6,267} \\
        \midrule
        \textbf{Image Magnification} & & \\
        \hspace{4mm} 10$\times$  & -- & 2,146 \\
        \hspace{4mm} 20$\times$  & -- & 4,121 \\
        \bottomrule
    \end{tabular}
\end{table}
Cases were categorized into two primary classes: a mycosis fungoides (MF) cohort and a heterogeneous non-MF 
cohort. 
As detailed in Table \ref{tab:dataset_distribution}, the non-MF cohort spans several 
lymphoproliferative skin disorders acting as clinical and histological mimics of MF.

The dataset was partitioned at the patient level into 329 training patients, 
67 validation patients, and 67 test patients. 
The acquired images were subdivided into patches for subsequent model training, validation, and testing. 
In particular, the 20$\times$ images were subdivided into 23,062 training patches, 4,728 validation 
patches, and 4,320 test patches, 
while the 10$\times$ images were subdivided into 73,824 training patches, 14,804 validation patches, 
and 12,848 test patches.

\subsection{Preprocessing}
All input images were provided in a \texttt{.TIF} format and acquired at a native resolution of $2560 \times 1920$ pixels. 
The images were hematoxylin-and-eosin (H\&E) stained and captured at fixed optical magnifications.
We used a sliding window approach to extract diagnostically relevant patches. 
To capture appropriate spatial context, the extraction parameters were tailored to the optical magnification:
\begin{itemize}
    \item \textbf{10$\times$ magnification:} $512 \times 512$ pixel patches with a stride of 256 pixels (50\% overlap).
    \item \textbf{20$\times$ magnification:} Native $1024 \times 1024$ patches 
    (with a stride of 512 pixels) were extracted to match the physical field of view of the $10\times$ 
    patches. \textit{These were subsequently downsampled to $512 \times 512$
     for network input.} Despite this digital resizing, initial capture through a 
     $20\times$ objective lens preserves superior optical resolution and cytological details.
\end{itemize}

\begin{figure}
    \centering
    \includegraphics[width=1\linewidth]{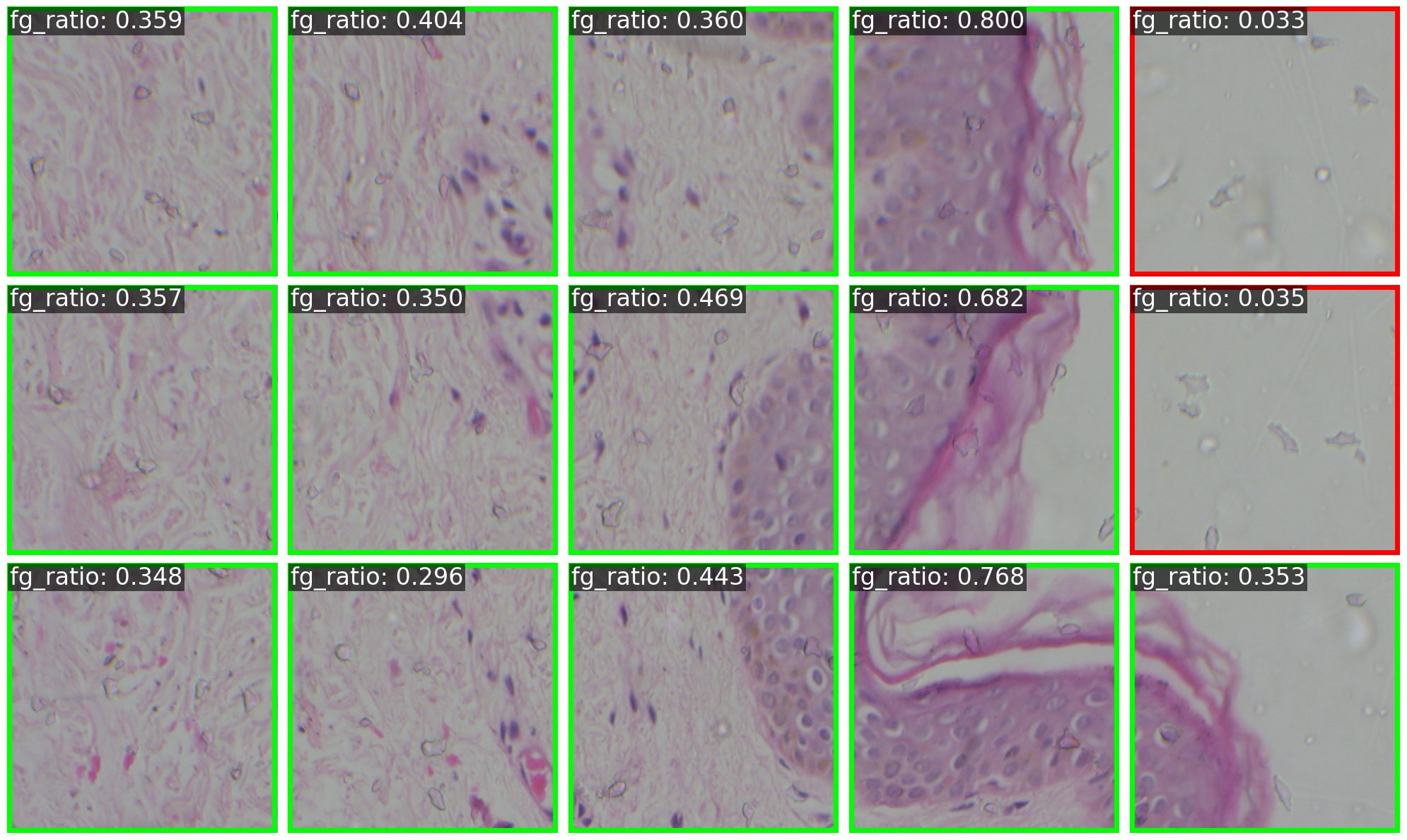}
    \caption{Different foreground ratios for patches of a sample TIF image}
    \label{fig:HSV_Ratio}
\end{figure}

To eliminate non-informative background prior to model training, RGB mean-intensity 
thresholding proved insufficient due to staining and illumination inconsistencies. 
Consequently, a robust saturation-based filtering strategy was adopted in the HSV color space, where patches 
with a foreground pixel ratio below 28.5\% were discarded (Figure \ref{fig:HSV_Ratio}), 
effectively suppressing background while preserving relevant tissue. 
To enhance generalization and mitigate class imbalance, the training set exclusively 
underwent spatial augmentations (random horizontal/vertical flips and limited rotations) for enhanced 
orientation-invariance, alongside photometric adjustments 
(brightness, contrast, saturation, and hue) for more staining variability. 
Finally, augmented patches were converted into tensors and normalized using ImageNet 
statistics.

\subsection{Model Architecture and Training Protocol}

\subsubsection{Dual-Magnification Architecture}
To capture both broad architectural distortion and granular cellular atypia, we deployed a 
two-branch architecture utilizing independent feature extractors. Both models leverage the 
\textbf{TF-EfficientNet-B3} backbone (with approximately 10.3M parameters) to distinguish between 
5 target classes (MF, PLEVA-PLC, Pseudolymphoma, T-cell dyscrasia, and B-cell Lymphoma). 
The configurations for each magnification level are summarized in Table \ref{tab:hyperparams}.

\begin{table}[htbp]
\caption{Model Configuration and Hyperparameters}
\label{tab:hyperparams}
\centering
\renewcommand{\arraystretch}{1.1}
\begin{tabular}{lcc}
\toprule
\textbf{Parameter} & \textbf{10$\times$ Model} & \textbf{20$\times$ Model} \\
\midrule
Base Architecture & TF-EfficientNet-B3 & TF-EfficientNet-B3 \\
Input Resolution & $512 \times 512$ & $512 \times 512$ \\
Dropout Rate & 0.40 & 0.40 \\
Inference Batch Size & 8 & 2 \\
Validation Metric & F$_2$-Score & F$_2$-Score \\
\bottomrule
\end{tabular}
\end{table}

\subsubsection{Training Strategy and Augmentation}
Both the 10$\times$ architectural and 20$\times$ cytological models were 
trained utilizing the AdamW optimizer with a static learning rate of 
$2 \times 10^{-4}$ and a weight decay of 0.01. To address the severe class 
imbalance inherent in the dataset, a dynamically weighted cross-entropy loss 
function was applied to both branches. Class penalty weights were calculated 
dynamically to be inversely proportional to their representation in the training 
distribution, forcing the network to prioritize rare cytological 
variants without requiring specialized oversampling techniques.

To enhance generalization, a shared spatial and color augmentation 
pipeline was applied during training. This pipeline included random 
horizontal and vertical flips, random rotations ($\pm15^\circ$), and 
color jittering (brightness, contrast, and saturation variations of $\pm20\%$, 
and hue variations of $\pm5\%$). Training leveraged automatic mixed precision (AMP) 
and gradient accumulation 
(8 steps for the 20$\times$ model, and 1 step for the 10$\times$ model) to optimize 
memory utilization and stabilize batch gradients.

\subsubsection{Metric Selection and Checkpointing}
Given the critical nature of cancer diagnosis, minimizing false negatives is paramount. 
Consequently, the checkpointing strategy was tailored to the specific role of each model. 
The 10$\times$ model checkpointing was governed by the F$_2$-score, which weighs recall (sensitivity) 
twice as heavily as precision:

\begin{equation}
    F_2 = (1 + 2^2) \cdot \frac{\text{Precision} \cdot \text{Recall}}{(2^2 \cdot \text{Precision}) + \text{Recall}}
    \label{eq:f2_score}
\end{equation}

To ensure the networks prioritized minority class detection over sheer accuracy, 
checkpointing for both the 10$\times$ and 20$\times$ models was strictly 
governed by the macroscopic \textbf{F2-Score evaluated on the validation set.} 
This metric was deliberately chosen to heavily penalize false negatives, 
maintaining a high sensitivity for critical cytological differentiation.

\subsection{Patient-Level Prediction and Decision Thresholding}
Model inference was initially performed at the patch level. To obtain clinically meaningful predictions at the patient level, patch-level probabilities corresponding to the same patient were aggregated using mean aggregation. Given a patient $p$ with $N_p$ patches and predicted probabilities $\{s_1, s_2, \ldots, s_{N_p}\}$, the patient-level score $S_p$ was computed as:
\begin{equation}
    S_p = \frac{1}{N_p} \sum_{i=1}^{N_p} s_i
\end{equation}

\subsubsection{Threshold Optimization via Youden's J Statistic}
Threshold analysis revealed fundamentally different probability distributions between the architectural (10$\times$) and 
cytological (20$\times$) models. While the 10$\times$ model was naturally balanced at a 0.5 decision boundary, 
the 20$\times$ model exhibited a severe conservative bias at the same threshold, yielding an unacceptable MF 
sensitivity of just 43.48\%. A systematic threshold sweep demonstrated that the 20$\times$ model's true 
discriminatory power was shifted heavily to the right of the probability axis. By applying 
Youden's J statistic ($J = \text{Sensitivity} + \text{Specificity} - 1$), an optimal threshold of 0.8351 was 
identified for the 20$\times$ model, which restored clinical viability by maximizing the overall accuracy to 83.58\% and 
elevating MF sensitivity to 89.13\% without degrading non-MF detection. This substantial discrepancy between the two 
models' operating points necessitated the piecewise-linear probability calibration applied in the subsequent 
late-fusion pipeline (Section~\ref{sec:fusion}).

\subsection{Multi-Magnification Fusion Strategy}\label{sec:fusion}

\subsubsection{Late Fusion Approach}
To integrate complementary diagnostic features captured at different scales, we employed a \textit{late fusion} strategy. Unlike early fusion, which concatenates feature vectors prior to classification, late fusion aggregates the probabilistic predictions of independent models. This approach allows each network to specialize in the specific histological patterns visible at its respective magnification—capturing architectural disarray at $10\times$ and cellular atypia at $20\times$—without the risk of feature-space contamination.

\subsubsection{Weighted Score Aggregation and Clinical Justification}
In clinical practice, dermatopathologists employ a multi-scale workflow: they initially scan tissue at low magnification to identify broad architectural abnormalities (e.g., epidermotropism) before zooming in to high magnification to assess individual cellular atypia. Our late-fusion approach mimics this workflow by integrating both perspectives.

Let $S_{p}^{10}$ and $S_{p}^{20}$ denote the mean-aggregated patient-level probabilities obtained from the $10\times$ and $20\times$ models, respectively. The final fused diagnostic score $S_{p}^{fused}$ was computed as a linear combination:
\begin{equation}
    S_{p}^{fused} = \alpha S_{p}^{10} + (1 - \alpha) S_{p}^{20}
\end{equation}
where $\alpha \in [0, 1]$ represents the fusion weight governing the contribution of the $10\times$ model.

\subsubsection{Optimization of the Fusion Weight}
The optimal weight $\alpha$ was determined empirically on the validation set. 
Unlike early experiments on smaller cohorts, the optimization on this expanded dataset consistently assigned a 
dominant weight to the $20\times$ model. The optimal balance was achieved at $\alpha = 0.20$, meaning the final 
diagnostic score relies 80\% on the $20\times$ cellular features and 20\% on the $10\times$ architectural context. 

Clinically, this is a sound outcome: while mycosis fungoides exhibits broad architectural 
distribution (e.g., epidermotropism) captured at $10\times$, the definitive presence of cellular 
atypia (such as hyperchromatic, cerebriform lymphocytes) resolved at $20\times$ proved highly 
discriminative in a larger, more diverse patient population. Consequently, $\alpha = 0.20$ was adopted for 
all subsequent evaluations, yielding an optimal sensitivity of 0.8913 and ROC-AUC of 0.8778.

\begin{figure}
    \centering
    \includegraphics[width=1\linewidth]{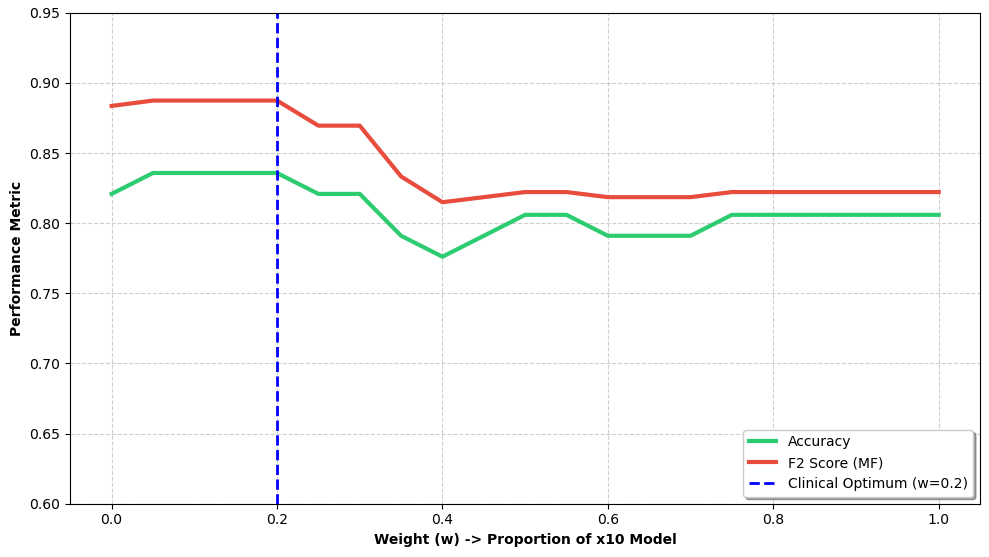}
    \caption{Optimization of multimodal late fusion weights. The graph illustrates the impact 
    of varying the 10$\times$ model's weight ($w$) on overall Accuracy and MF-specific F2-Score. 
    The clinical optimum was identified at $w=0.2$, demonstrating that a fusion ratio 
    heavily weighted toward 20$\times$ cytological features (80\%) augmented by 10$\times$ 
    architectural context (20\%) optimally balances high accuracy with the strict 
    sensitivity required for MF screening.}
    \label{fig:fusion_analysis}
\end{figure}

\begin{table}[htbp]
    \centering
    \caption{Univariate statistical analysis of clinical features ranked by $p$-value. Features with $p < 0.05$ were selected; Nodule was retained despite $p > 0.05$ for its staging value (see text).}
    \label{tab:clinical_univariate}
    \begin{tabular}{@{}lllc@{}}
        \toprule
        \textbf{Clinical Feature} & \textbf{Data Type} & \textbf{Statistical Test} & \textbf{$p$-value} \\
        \midrule
        \multicolumn{4}{c}{\textit{Selected Features (Significant)}} \\
        \midrule
        Macules & Categorical & Chi-Squared & $<0.001$ \\
        Biopsy 1 Morphology & Categorical & Chi-Squared & $<0.001$ \\
        Papules & Categorical & Chi-Squared & $<0.001$ \\
        Age & Continuous & Mann-Whitney U & $<0.001$ \\
        Color & Categorical & Chi-Squared & $<0.001$ \\
        Patch & Categorical & Chi-Squared & $<0.001$ \\
        Biopsy 2 Morphology & Categorical & Chi-Squared & $<0.001$ \\
        Visit Type & Categorical & Chi-Squared & $<0.001$ \\
        Duration (Months) & Continuous & Mann-Whitney U & $<0.001$ \\
        Plaque & Categorical & Chi-Squared & $<0.001$ \\
        Scales & Categorical & Chi-Squared & $<0.001$ \\
        Nodule$^{\dagger}$ & Categorical & Chi-Squared & $0.2007$ \\
        Site: Head/Neck & Categorical & Chi-Squared & $0.0001$ \\
        Site: Lower Limbs & Categorical & Chi-Squared & $0.0090$ \\
        Disease Course & Categorical & Chi-Squared & $0.0100$ \\
        Biopsy 2 Site & Categorical & Chi-Squared & $0.0397$ \\
        \midrule
        \multicolumn{4}{c}{\textit{Excluded Features (Not Significant)}} \\
        \midrule
        Biopsy 1 Site & Categorical & Chi-Squared & $0.0718$ \\
        Sex & Categorical & Chi-Squared & $0.4065$ \\
        Symptomatic & Categorical & Chi-Squared & $0.6562$ \\
        Site: Trunk & Categorical & Chi-Squared & $0.7829$ \\
        Site: Upper Limbs & Categorical & Chi-Squared & $0.8019$ \\
        Site: Buttocks & Categorical & Chi-Squared & $0.9292$ \\
        \bottomrule
    \end{tabular}
\end{table}

\subsection{Clinical Features Classifier}
To complement the histopathological image analysis, a machine learning module was developed to evaluate structured patient metadata drawn from a separate clinical cohort of 499 patients (353 MF, 146 Non-MF). Within the MF cohort, 326 patients presented at the patch-plaque stage and 27 at the tumor stage, enabling a hierarchical two-task prediction pipeline.

Univariate analysis of 22 clinical variables—using the Mann-Whitney U test for skewed continuous data and 
the chi-squared test for categorical data—identified 16 significant features ($p < 0.05$) for model training 
(Table~\ref{tab:clinical_univariate}). These encompass demographics (age), disease history 
(duration, visit type, course), anatomic distribution (head/neck, lower limbs), and lesion morphology 
(color, macules, patches, papules, plaques, scales, biopsy findings).

Notably, although the \textit{Nodule} feature is statistically insignificant for binary MF diagnosis 
($p = 0.2007$), it was deliberately retained. Nodules serve as a near-deterministic marker for tumor staging, 
making this feature's inclusion essential for the secondary intra-MF staging task.

The clinical classifier operates as a sequential two-task pipeline:
\begin{enumerate}
    \item \textbf{Task~1 --- Diagnosis (MF vs.\ Non-MF):} All 499 patient records are used to predict whether a patient has MF or a non-MF lymphoproliferative condition. All 16 selected features are utilized.
    \item \textbf{Task~2 --- MF Staging (Patch--Plaque vs.\ Tumor):} For patients classified as MF in Task~1, a second model predicts the clinical stage. This model operates on the 353 MF patient records and leverages the same feature set, where the \textit{Nodule} feature now becomes highly discriminative.
\end{enumerate}

During preprocessing, missing continuous values were filled in using median imputation, 
and categorical variables underwent label encoding. Positive class weights were adjusted to 
mitigate class imbalance. Three distinct algorithms were systematically compared using 5-fold 
stratified cross-validation: XGBoost, random forests (with 300 estimators and a max depth of 6), and logistic 
Regression. Random forests were selected as the primary model for both tasks based on overall 
cross-validation performance.

\section{Results}

\subsection{Deep Learning Image Analysis Results}
The proposed deep learning framework was evaluated on a held-out test set of 67 patients, comprising 12,848 patches at 10$\times$ magnification and 4,320 patches at 20$\times$ magnification. As summarized in Table \ref{tab:results}, the system achieved a maximum accuracy of 83.58\% and an F$_2$-score of 0.8874 using the late-fusion model.

\begin{table}[htbp]
\caption{Quantitative Performance Summary (Test Set, N=67 Patients)}
\label{tab:results}
\centering
\renewcommand{\arraystretch}{1.1}
\setlength{\tabcolsep}{4pt} 
\begin{tabular}{l c c c}
\toprule
 & \textbf{10$\times$ Model} & \textbf{20$\times$ Model} & \textbf{Late Fusion ($\alpha=0.20$)} \\
\midrule
Accuracy & 0.8060 & 0.8358 & \textbf{0.8358} \\
F$_2$-Score & 0.8222 & \textbf{0.8874} & \textbf{0.8874} \\
Sensitivity & 0.8043 & \textbf{0.8913} & \textbf{0.8913} \\
Specificity & 0.8043 & \textbf{0.8913} & \textbf{0.8913} \\
ROC-AUC & 0.8685 & 0.8509 & \textbf{0.8778} \\
\bottomrule
\end{tabular}
\end{table}
\subsection{Patient-Level Aggregation Benefits}
\begin{figure}[!t]
    \centering
    \includegraphics[width=1\linewidth]{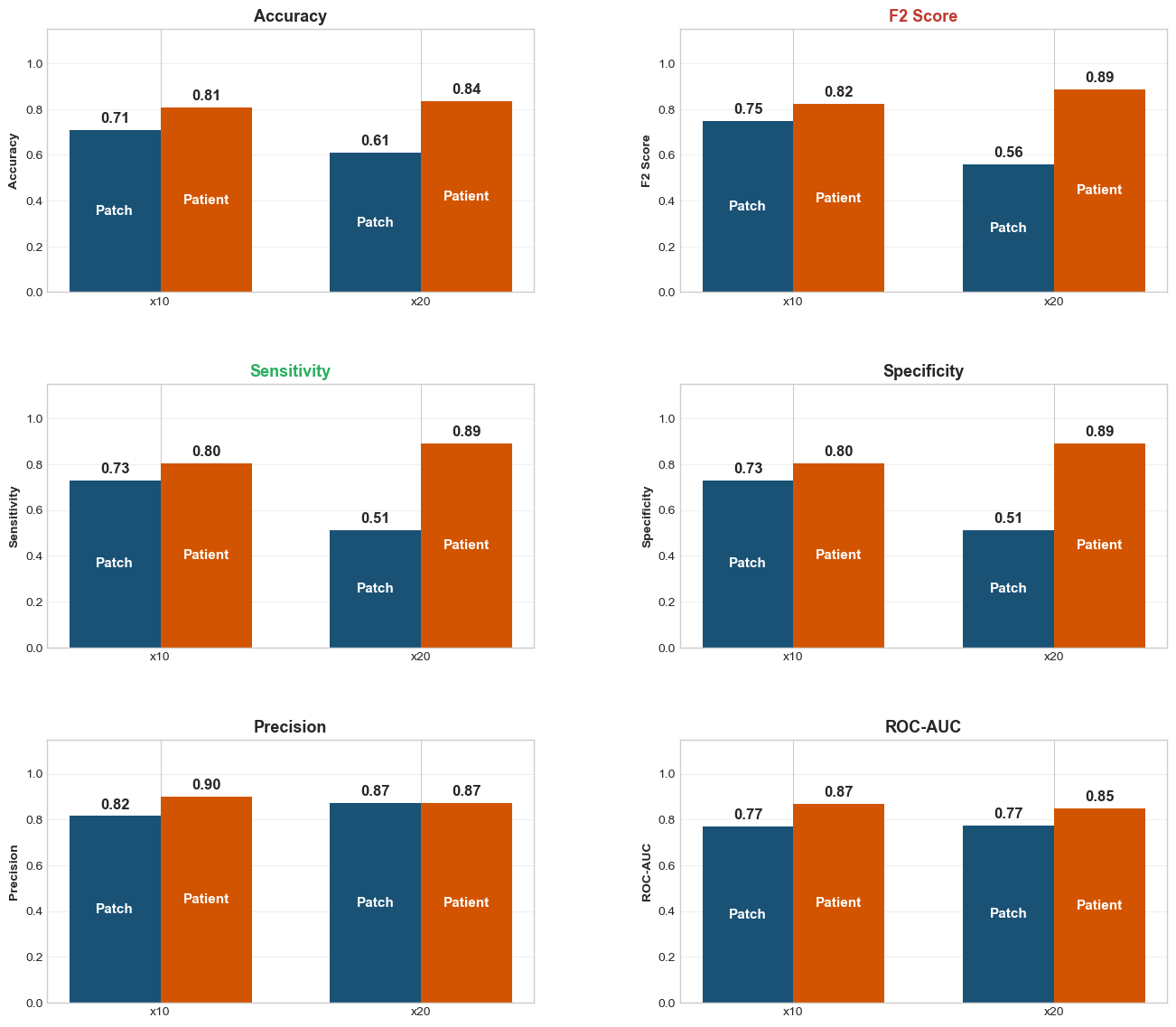}
    \caption{A side-by-side comparison of patch-level (dark blue) vs. patient-level (dark orange) metrics across both magnifications.}
    \label{fig:comprehensiveMetrics}
\end{figure}

A key finding of this study is the significant performance gain achieved through patient-level 
aggregation. As illustrated in 
(Fig. \ref{fig:comprehensiveMetrics}), patient-level consensus outperformed patch-level 
predictions on every metric.

The late fusion model maintains the exceptional sensitivity (0.8913) and F$_2$-score (0.8874) of the 20$\times$ patient-level model, while achieving a robust ROC-AUC (0.8778) and accuracy (83.58\%). By relying heavily on the 20$\times$ model's cytological precision ($\alpha=0.20$), the late fusion model effectively maximizes the detection of true positives, ensuring that ambiguous but potentially malignant cases are not missed.

\subsection{Sensitivity and Clinical Relevance}
Given the clinical imperative to minimize missed diagnoses, we prioritized the F$_2$-score (which emphasizes recall).
The 20$\times$ model achieved a superior F$_2$-score of 0.8874 compared to 0.8222 for the 10$\times$ model. This confirms that while low-power architectural changes are helpful, the high-magnification cellular details—such as hyperchromatic and cerebriform nuclei—are the most critical and discriminative indicators of MF.

\subsection{Clinical Feature Classification Results}
The clinical machine learning module demonstrated exceptional predictive power across both tasks.

\subsubsection{Task~1: Diagnosis (MF vs.\ Non-MF)}
For 5-fold stratified cross-validation on the 499 patient records, the random forest model emerged as the dominant classifier. As detailed in Table~\ref{tab:clinical_task1}, random forest achieved the highest accuracy (96.6\%), precision (94.8\%), F$_2$-score (0.939), and ROC-AUC (0.995). While XGBoost exhibited marginally higher sensitivity (94.5\% vs.\ 93.8\%), the random forest model's superior precision and overall discrimination capability established it as the best clinical model. Logistic regression served as a strong linear baseline but was outperformed by the tree-based ensembles, highlighting the non-linear feature interactions within clinical presentations of MF.

\begin{table}[htbp]
\caption{Task~1: MF vs.\ Non-MF Diagnosis (5-Fold Stratified CV)}
\label{tab:clinical_task1}
\centering
\renewcommand{\arraystretch}{1.1}
\setlength{\tabcolsep}{4pt}
\begin{tabular}{l c c c}
\toprule
\textbf{Metric} & \textbf{XGBoost} & \textbf{Random Forest} & \textbf{Logistic Reg.} \\
\midrule
Accuracy    & 0.956              & \textbf{0.966}          & 0.950 \\
Precision   & 0.908              & \textbf{0.948}          & 0.915 \\
Sensitivity & \textbf{0.945}     & 0.938                   & 0.918 \\
F$_2$-Score & 0.937              & \textbf{0.939}          & 0.916 \\
ROC-AUC     & 0.994              & \textbf{0.995}          & 0.985 \\
\bottomrule
\end{tabular}
\end{table}

\begin{figure}[!t]
    \centering
    \includegraphics[width=0.8\linewidth]{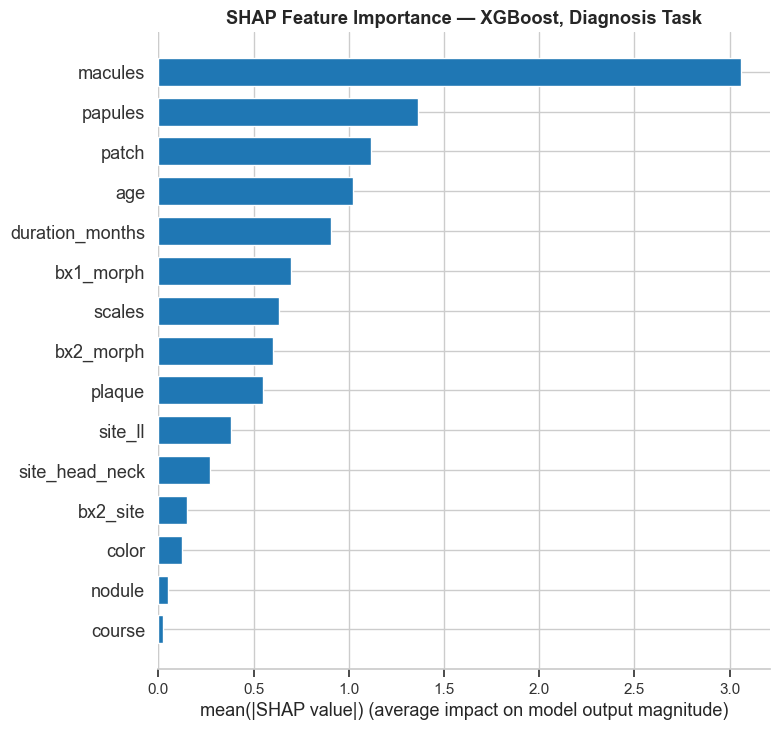}
    \caption{SHAP feature importance for Task~1 (MF vs.\ Non-MF). Features ranked by the mean absolute SHAP value.}
    \label{fig:shap_importance_task1}
\end{figure}

\begin{figure}[!t]
    \centering
    \includegraphics[width=0.8\linewidth]{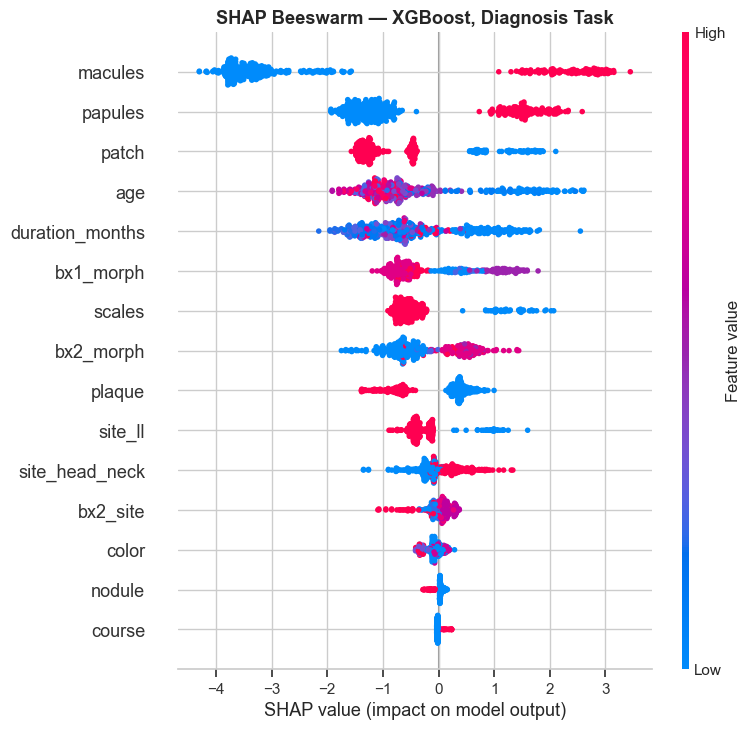}
    \caption{SHAP beeswarm plot for Task~1. Each dot represents a single patient; color indicates feature value (red = high, blue = low), and horizontal position shows the SHAP value (impact on prediction).}
    \label{fig:shap_beeswarm_task1}
\end{figure}

SHAP (SHapley Additive exPlanations) analysis was performed to investigate the key contributors to the performance of the
random forest diagnosis model. As shown in Fig.~\ref{fig:shap_importance_task1}, 
the top contributing features include macules, biopsy morphology, age, and disease 
duration—providing clinically interpretable explanations for individual predictions.
The corresponding SHAP beeswarm plot (Fig.~\ref{fig:shap_beeswarm_task1}) reveals 
the directional impact of each feature on the model's output.

\subsubsection{Task~2: MF Staging (Patch--Plaque vs.\ Tumor)}
For intra-MF staging on the 353 MF patient records, all three models achieved high accuracy ($>$95\%), reflecting the strong discriminative power of lesion morphology features at this stage. As shown in Table~\ref{tab:clinical_task2}, random forest again achieved the highest accuracy (97.7\%) and F$_2$-Score (0.855), while XGBoost maintained competitive performance. Logistic regression exhibited a marked decline in sensitivity (52.7\%) and F$_2$-Score (0.532) for this task, underscoring the importance of non-linear models for capturing the complex interactions between staging features.

\begin{table}[htbp]
\caption{Task~2: MF Staging Results --- Patch--Plaque vs.\ Tumor (5-Fold Stratified CV)}
\label{tab:clinical_task2}
\centering
\renewcommand{\arraystretch}{1.1}
\setlength{\tabcolsep}{4pt}
\begin{tabular}{l c c c}
\toprule
\textbf{Metric} & \textbf{XGBoost} & \textbf{Random Forest} & \textbf{Logistic Reg.} \\
\midrule
Accuracy    & 0.972              & \textbf{0.977}          & 0.952 \\
Precision   & 0.828              & \textbf{0.870}          & 0.700 \\
Sensitivity & \textbf{0.860}     & \textbf{0.860}          & 0.527 \\
F$_2$-Score & 0.843              & \textbf{0.855}          & 0.532 \\
ROC-AUC     & \textbf{0.990}     & 0.987                   & 0.968 \\
\bottomrule
\end{tabular}
\end{table}

\begin{figure}[!t]
    \centering
    \includegraphics[width=1\linewidth]{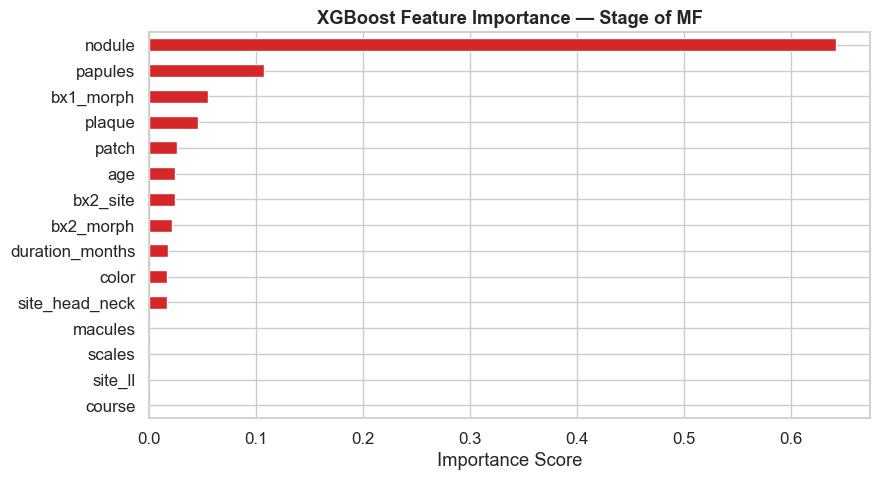}
    \caption{Feature importance for Task~2 (Stage of MF). The \textit{Nodule} feature dominates, confirming its near-deterministic role in identifying the tumor stage.}
    \label{fig:fi_stage}
\end{figure}

Feature importance analysis for the staging task (Fig.~\ref{fig:fi_stage}) reveals that the \textit{Nodule} feature is the single most important predictor, confirming the clinical rationale for its retention despite its lack of statistical significance in the diagnosis task. This finding aligns with the well-established dermatological principle that the presence of nodules/tumors is a hallmark of advanced-stage MF.

\section{Discussion}
\subsection{Cellular Detail within Architectural Context}
A key finding of this study is the superior performance of the 20$\times$ model (F$_2$=0.8874) compared to the 10$\times$ model (F$_2$=0.8222). In histopathology, higher magnification typically yields better resolution of cellular atypia, such as hyperchromatic and irregular cerebriform nuclei, which are hallmark cytological features of mycosis fungoides. Our empirical results strongly support this established dermatopathologic principle, demonstrating that high-resolution cellular detail is the primary driver of accurate algorithmic diagnosis.

However, mycosis fungoides is also defined by architectural patterns, such as epidermotropism and the formation of Pautrier's microabscesses, which are more readily apparent at lower magnifications. Our dual-scale late-fusion approach successfully mimics the workflow of a human pathologist by combining both perspectives. By assigning 80\% of the decision weight to the 20$\times$ cytological model and 20\% to the 10$\times$ architectural model, the framework explicitly prioritizes definitive cellular evidence while still benefiting from the broader architectural context, resulting in a highly sensitive and reliable diagnostic tool.

\subsection{The Necessity of Whole-Slide Context}
The success of patient-level aggregation highlights the diffuse nature of MF: features are spread subtly across the tissue rather than concentrated in isolated hotspots. Mean aggregation integrates the probabilistic signal from all patches, acting as a digital analog to a pathologist scanning the entire slide before rendering a diagnosis.

\subsection{Benchmarking and Clinical Reality}
Diagnosing MF solely from histopathological images is a simplified proxy for clinical reality, where pathologists synthesize visual data with patient history and clinical presentation. Despite this limitation, our framework is highly competitive with recent state-of-the-art benchmarks (summarized in Table \ref{tab:sota_comparison}). 

\begin{table*}[htbp]
\caption{Comparison of Recent AI Frameworks for Mycosis Fungoides (MF) Diagnosis}
\label{tab:sota_comparison}
\centering
\renewcommand{\arraystretch}{1.3}
\begin{tabular}{p{2.5cm} p{3.5cm} p{3.5cm} p{2.5cm} p{3.5cm}}
\toprule
\textbf{Study} & \textbf{Data Modality} & \textbf{Clinical Task} & \textbf{Reported Metric} & \textbf{Deployment Requirement} \\
\midrule
Doeleman et al. (2024) \cite{Doeleman2024-MF} & WSI (H\&E) & Early MF vs. Benign & Mean AUC: 0.827 & High (Digital WSI Scanner) \\
Zhao et al. (2026) \cite{10.1093/bjd/ljag129} & WSI + Clinical Data & MF vs. Benign Inflammatory & Macro-AUC: $>$0.85 & High (Digital WSI Scanner) \\
Liu et al. (2025) \cite{10.1093/bjd/ljaf212} & Dermoscopy + Clinical Images & Early MF vs. Inflammatory & Accuracy: 82.9\%* & Medium (Dermoscopy equipment) \\
Ghosh et al. (2026) \cite{10.1093/bjd/ljaf345} & Non-Linear Optical Microscopy & Detection of Epidermotropism & Qualitative & Very High (NLOM infrastructure) \\
\textbf{Ours} & \textbf{Fixed-Magnification (10$\times$, 20$\times$) H\&E} & \textbf{MF vs. Non-MF} & \textbf{Accuracy: 83.58\%} & \textbf{Low (Standard Optical Microscope)} \\
\bottomrule
\multicolumn{5}{l}{\footnotesize *Performance reported for the Dermatologist + AI assisted group.}
\end{tabular}
\end{table*}

\subsection{Comparison with State-of-the-Art}
Recent AI systems for MF diagnosis demonstrate high efficacy but heavily rely on expensive infrastructure. 
For instance, as seen in table \ref{tab:sota_comparison}, Doeleman et al. \cite{Doeleman2024-MF} (mean AUC 0.827) and Zhao et al. \cite{10.1093/bjd/ljag129} 
(Macro-AUC $>0.85$) utilize computationally heavy Whole-Slide Images (WSIs). Similarly, Ghosh et al. 
\cite{10.1093/bjd/ljaf345} require advanced Non-Linear Optical Microscopy (NLOM), while Liu et al. 
\cite{10.1093/bjd/ljaf212} rely on specialized dermoscopy. 

These hardware requirements limit deployment in resource-constrained clinical settings. 
In contrast, our dual-scale late-fusion framework operates directly on standard, fixed-magnification 
($10\times$ and $20\times$) optical microscopy images. By achieving a patient-level accuracy of 83.58\%, 
sensitivity of 89.13\%, and an ROC-AUC of 0.797, our method competitively matches these benchmarks without 
requiring whole-slide digitization, providing a highly accessible diagnostic alternative for public hospital 
laboratories.
\section{Conclusion}
This work presents a comprehensive, multimodal machine learning framework for automated MF screening. Our key contributions are:
\begin{enumerate}
    \item \textbf{Cytological detail dominance:} The 20$\times$ model's capacity to resolve fine cellular atypia significantly outperforms the broader 10$\times$ view. By allocating 80\% fusion weight to the 20$\times$ model, the late-fusion ensemble achieves an 89.13\% sensitivity and an 83.58\% accuracy, showing that high magnification is critical for accurate differential diagnosis.
    \item \textbf{Complementary clinical AI:} A random forest classifier trained on 16 structured clinical features independently achieved 93.8\% sensitivity and 96.6\% accuracy, demonstrating that algorithmic analysis of patient metadata is a highly reliable parallel modality to MF histopathology.
    \item \textbf{Multi-scale patient-level fusion:} The late-fusion protocol systematically combines cytological detail (20$\times$) with architectural context (10$\times$) via patient-level mean aggregation, achieving optimal screening performance.
\end{enumerate}
\section{Future Work}
To evolve this framework into a robust clinical decision-support system, future research will focus on two key areas. 
First, we aim to develop end-to-end multimodal architectures that utilize cross-attention mechanisms to directly integrate 
histopathological features with clinical tabular data, moving beyond independent models to achieve superior unified performance. 
Second, to ensure generalizability across diverse laboratory settings, subsequent iterations will incorporate rigorous 
stain normalization techniques (e.g., Macenko \cite{macenko2009} or Vahadane \cite{vahadane2016_stainnorm}) to mitigate the 
confounding effects of H\&E color variations.

\bibliographystyle{IEEEtran}
\bibliography{references}

\end{document}